\documentclass[10pt,twocolumn,letterpaper]{article}

\usepackage{wacv}              %

\usepackage[pagebackref,breaklinks,colorlinks]{hyperref}

\usepackage{graphicx}
\usepackage{booktabs}
\usepackage{amsmath}
\usepackage{amssymb}
\usepackage{mathtools}
\usepackage{makecell}
\usepackage{pifont}
\usepackage[accsupp]{axessibility}  %
\usepackage{orcidlink}
\usepackage{float}
\usepackage{amsfonts}
\usepackage{wrapfig}

\usepackage[capitalize]{cleveref}
\crefname{section}{Sec.}{Secs.}
\Crefname{section}{Section}{Sections}
\Crefname{table}{Table}{Tables}
\crefname{table}{Tab.}{Tabs.}

\usepackage{multirow}
\usepackage{color, colortbl}

\definecolor{Gray}{gray}{0.9}
\definecolor{lightgray}{gray}{0.5}
\definecolor{verylightgray}{gray}{0.7}
\definecolor{veryverylightgray}{gray}{0.9}

\definecolor{darkgreen}{rgb}{0, 0.4, 0}
\definecolor{darkred}{rgb}{0.7, 0, 0}
\definecolor{darkblue}{rgb}{0.0, 0.0, 0.7}

\usepackage{amsmath,amsfonts,bm,physics}

\newcommand{\pstud}{q_s}
\newcommand{\pteach}{q_t}

\def\rvn{{\mathbf{n}}}

\def\rvx{{\mathbf{x}}}
\def\rvy{{\mathbf{y}}}
\def\rvz{{\mathbf{z}}}

\DeclareMathAlphabet{\mathsfit}{\encodingdefault}{\sfdefault}{m}{sl}
\SetMathAlphabet{\mathsfit}{bold}{\encodingdefault}{\sfdefault}{bx}{n}

\DeclareMathOperator*{\argmax}{arg\,max}

\def\wacvPaperID{169} %
\def\confName{WACV}
\def\confYear{2025}

\begin{document}

\title{Rethinking cluster-conditioned diffusion models for label-free image synthesis}

\author{Nikolas Adaloglou\\
Heinrich Heine University of Dusseldorf\\
{\tt\small adaloglo@hhu.de}
\and
Tim Kaiser\\
Heinrich Heine University of Dusseldorf\\
{\tt\small tikai103@hhu.de}
\and
Felix Michels\\
Heinrich Heine University of Dusseldorf\\
{\tt\small felix.michels@hhu.de}
\and
Markus Kollmann\\
Heinrich Heine University of Dusseldorf\\
{\tt\small markus.kollmann@hhu.de}}
\maketitle

\begin{abstract}
Diffusion-based image generation models can enhance image quality when conditioned on ground truth labels. Here, we conduct a comprehensive experimental study on image-level conditioning for diffusion models using cluster assignments. We investigate how individual clustering determinants, such as the number of clusters and the clustering method, impact image synthesis across three different datasets. Given the optimal number of clusters with respect to image synthesis, we show that cluster-conditioning can achieve state-of-the-art performance, with an FID of 1.67 for CIFAR10 and 2.17 for CIFAR100, along with a strong increase in training sample efficiency. We further propose a novel empirical method to estimate an upper bound for the optimal number of clusters. Unlike existing approaches, we find no significant association between clustering performance and the corresponding cluster-conditional FID scores. Code is available at \url{https://github.com/HHU-MMBS/cedm-official-wavc2025} 
\end{abstract}

\section{Introduction}
Diffusion models have enabled significant progress in many visual generative tasks, such as image synthesis \cite{ho2020ddpm,edm, diffusion_2023_transformers} and manipulation \cite{yang2023paint}. Conditioning diffusion models on human-annotated data is today's standard practice as it significantly improves the image fidelity \cite{dhariwal2021diffusion,batzolis2021conditional_diffusion,bao2022why_cond}. Image-level conditioning is typically realized by using associated text captions \cite{saharia2022photorealistic, sheynin2022knn_diffusion_text} or class labels, if available \cite{ho2020ddpm,nichol2021improved_ddpm}. Nonetheless, human-annotated labels are costly and often contain inaccuracies \cite{beyer2020done_imagenet,barz2020cifar_duplicates}, while publicly available image-text pairs can be non-descriptive \cite{dalle3}. 

\begin{figure}
    \centering  \includegraphics[width=0.8\columnwidth]{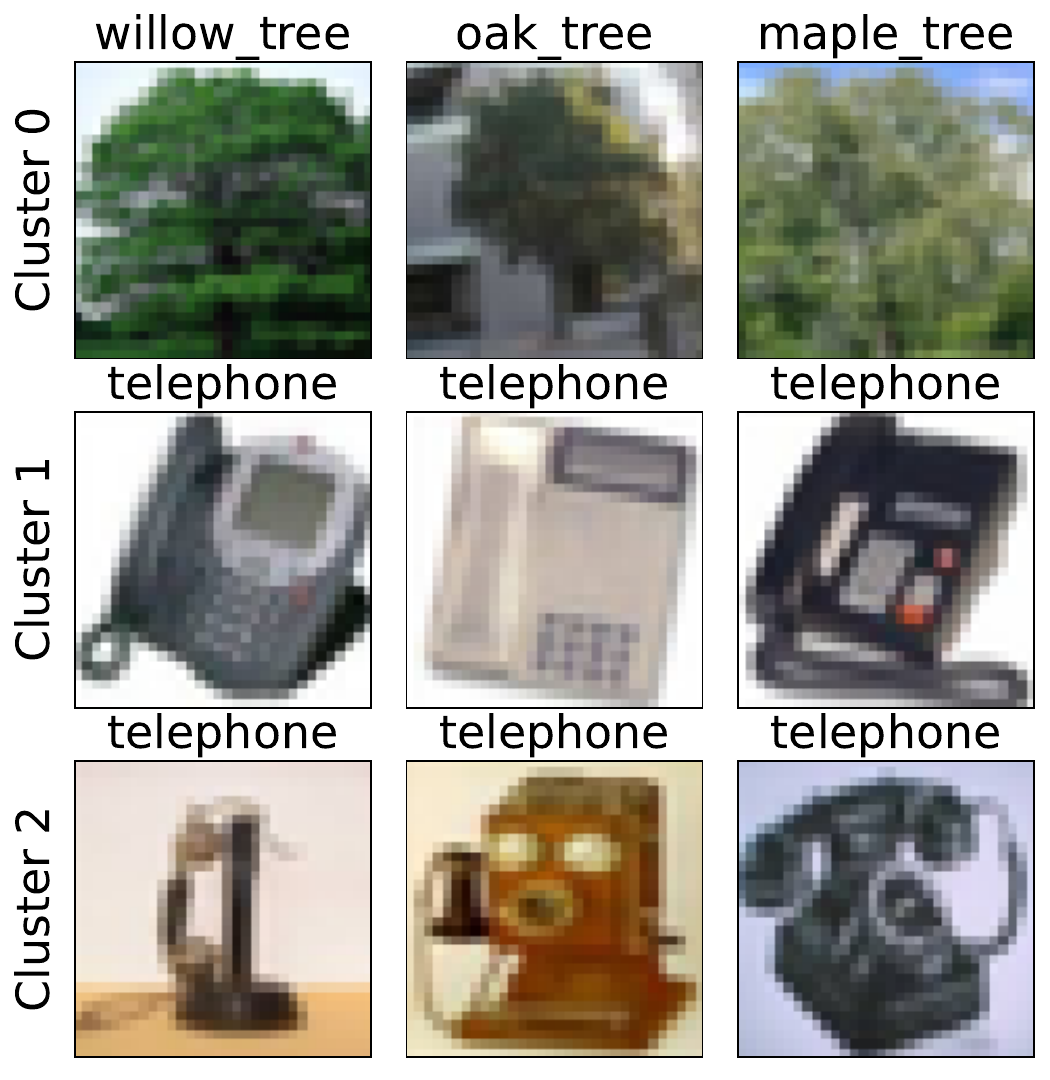}
  \caption{An ideal image-level conditioning should group images based on shared patterns, shown in the same row, which do not always align with human labels, indicated above each image (CIFAR100\cite{cifar} samples).}
    \label{fig:intro_fig}
\end{figure}

Large, human-annotated datasets are prohibitively costly \cite{radford2021clip} and hard to obtain in a plethora of real-life applications such as medical image synthesis \cite{kazerouni2022diffusion}. {Human annotation is subject to the label collection procedure and can have varying annotation hierarchies \cite{hollink2019fruit,beyer2020we,adaloglou2024scaling}. An ideal conditioning signal would be based solely on shared characteristics corresponding to general visual concepts, as illustrated in \cref{fig:intro_fig}. To provide a ground truth (GT) label-free alternative that is applicable to unlabelled datasets}, we focus on cluster-based conditioning on diffusion models. Image clustering refers to algorithmically assigning a semantic group to an image, called a cluster, given an \textit{a priori} number of such groups \cite{deep2018_cluster,chang2017deep_cluster,ji2019invariant_cluster}. 

Maximizing cluster alignment with ground truth (\emph{GT}) labels does not guarantee optimal cluster-conditional image generation. To create visually distinguishable groups (\cref{fig:intro_fig}), a clustering algorithm would sometimes group images into more fine-grained groups than the GT labels, i.e., \ sub-grouping analog and digital devices that map to the same label or merge similarly looking images with different labels (merging different types of trees).

Deep image clustering has recently seen significant progress \cite{xie2016unsupervised,sela,sscn}, especially when utilizing off-the-shelf features learned from large-scale pre-training \cite{scan,temi,adaloglou2024scaling}.  Currently, the relationship between clustering performance and cluster-conditional image synthesis performance remains unclear for several reasons. First, clustering metrics capture the alignment to GT labels, whereas generative models may benefit more from visually distinguishable groups \cite{hu2023self}. Second, clustering metrics cannot be computed for unlabeled datasets. Third, existing metrics are not suitable for fairly evaluating cluster assignments with varying numbers of clusters. Therefore, they cannot help to determine the optimal cluster granularity w.r.t.\ generative performance for an arbitrary dataset. This aspect has not been sufficiently explored to date \cite{cvpr2020gans_clusters,bao2022why_cond}.

Existing cluster-conditioned generative approaches \cite{cvpr2020gans_clusters,bao2022why_cond,hu2023self} often adopt simple generative baselines \cite{ho2020ddpm} and have mainly investigated $k$-means clustering on balanced classification datasets. The adoption of $k$-means is justified due to its ability to work with off-the-shelf feature extractors and its non-parametric nature, apart from setting the number of clusters. Conversely, it is well-established that $k$-means is suboptimal for image clustering because its cluster assignments are highly imbalanced \cite{scan, hu2023self}. {While cluster-conditioned approaches are sensitive to the number of clusters \cite{bao2022why_cond}, there is currently no method to predetermine this number. Previous methods have overlooked this \cite{cvpr2020gans_clusters,bao2022why_cond,hu2023self}, significantly hindering their applicability to unlabeled datasets. Recent feature-based image clustering methods have demonstrated superior performance compared to $k$-means \cite{scan, temi}, and their efficacy in conditional image generation is yet to be investigated. For the above reasons, cluster-conditioning methods underperform GT conditional ones to date.}

In this paper, we systematically study clustering as a label-free condition for diffusion models. We demonstrate that optimal cluster granularity achieves state-of-the-art FID scores (1.67 for CIFAR10 and 2.17 for CIFAR100) while enhancing training sample efficiency. We propose a computationally efficient method to derive an upper cluster bound using feature-based clustering, which narrows the search space for optimal image synthesis performance. We validate this upper bound across three datasets and two generative metrics. Finally, we find no significant association between clustering performance and cluster-conditional image synthesis performance across various clustering methods, performance metrics, and pre-trained models.

\section{Related Work}
\subsection{Conditional generative models}
Early attempts on controllable image synthesis adopted generative adversarial networks (GANs) \cite{mirza2014conditional}. By conditioning both the generator and discriminator on human labels, GANs can produce images for a specific GT label, even at the scale of ImageNet \cite{brock2019large}. Recently, diffusion models (DMs) have emerged as an expressive and flexible category of generative models \cite{sohl2015deep, song2020denoising,ho2022cfg,rombach2021stable_diffusion, karras2023edm2}. Internal guidance-based methods \cite{dhariwal2021diffusion_classifier_guid,ho2022cfg,diffusion-self-attention2023iccv,hu2023guided_diffusion_cluster_new_paper} further improved the flexibility and visual fidelity of DMs during sampling such as classifier and classifier-free guidance. DMs exhibit an enormous number of label-conditional variants \cite{ho2020ddpm,kingma2021variational,edm, nichol2021improved_ddpm,peebles2023dit}. Recently, Stein et al. \cite{stein2023exposing} proposed the adoption of Fréchet DINOv2 \cite{oquab2023dinov2} distance as it aligns better with human preferences and demonstrated that DMs achieve the highest perceptual realism.

\subsection{Alternative conditioning of generative models}
Unlike internal guidance \cite{diffusion-self-attention2023iccv,ho2022cfg,hu2023guided_diffusion_cluster_new_paper}, external conditioning signals are computed from the training set using additional models, which is commonly referred to as \emph{self-conditioning}. Self-conditioning signals can be roughly divided into image-level and sub-image-level. Image-level conditioning refers to a single condition for all pixels in an image, such as cluster assignments or text captions \cite{rombach2021stable_diffusion}. Sub-image-level conditions refer to specific parts or regions of an image \cite{hu2023self,bansal2023universal}. For instance, Hu et al. \cite{hu2023self} extend DMs to incorporate conditioning of bounding boxes or segmentation masks. 

\noindent\textbf{Feature-based conditioning.} Using image or text features can also provide an informative conditioning signal \cite{bordes2022ssl_knn_feats_diffusion,sheynin2022knn_diff,meta2023gen_representations,ramesh2022hierarchical}. Bordes et al. \cite{bordes2022ssl_knn_feats_diffusion} condition DMs directly on image representations (\emph{oracle features from real images}) and reveal their impact on visual fidelity. Instance-conditioned GAN combines GT labels with features from each image's NN set \cite{casanova2021icgan}. Ramesh et al. \cite{ramesh2022hierarchical} leverage the zero-shot capability of CLIP to condition DMs on language-guided image representations to improve diversity. Zhou et al. provide a language-free framework \cite{zhou2022towards} that directly generates text features for images using CLIP. Still, CLIP models require an \textit{a priori} set of descriptive candidate captions, such as label names.

\noindent\textbf{Cluster-based conditioning.} Similar to label conditioning, conditioning on cluster assignments facilitates DMs by allowing them to specialize on a distinct set of shared visual features \cite{hu2023self,bao2022why_cond,cvpr2020gans_clusters}. In \cite{bao2022why_cond}, Bao et al. computed the $k$-means clusters offline using contrastive learned features on the training data. However, their approach could not outperform the label-conditioned models. The first cluster-based approach that achieves competitive performance compared to GT labels leverages pre-trained feature extractors \cite{hu2023self}, which is the closest to our work. In \cite{hu2023self}, the authors attempt to weak establish a correlation between clustering and cluster-conditional generative performance for different self-supervised models. The observed correlation cannot be tested on unlabeled datasets, and its sensitivity to the number of clusters is unknown. While it was demonstrated that DINO \cite{dino} provides the most informative $k$-means clusters \cite{hu2023self}, their method lacks a strategy for choosing the cluster granularity for unlabeled datasets.

\section{Method}
\subsection{Notations and prerequisites} \label{sec:prerequisites}
We consider the frequent scenario where we have access to an \textbf{unlabeled} dataset $D$ and a pre-trained feature extractor $g(\cdot)$. We denote the number of ground truth (GT) labels as $C_{GT}$, which we assume to be unknown. To distinguish cluster-based conditioning from the image clustering task, we denote the number of \emph{visual groups} $C_{V}$ of $D$ as the optimal number of clusters w.r.t.\ image synthesis (e.g.\ as measured by FID \cite{heusel2017fid}). We adopt the diffusion model approach of Karras et al. (EDM\ \cite{edm}) as a baseline, which introduced various improvements to the standard DDPM \cite{ho2020ddpm}. We provide an overview of EDM in the supplementary material. Although we focus on EDM throughout this work, our method can be applied to any conditional generative model.

\newcommand{\temiloss}{\mathcal{L}_\text{\sc{temi}}}
\noindent\textbf{TEMI clustering.}
Given an \textit{a priori} determined number of clusters $C$ and a feature extractor $g(.)$, TEMI \cite{temi} first mines the $m$ nearest neighbors (NN) of all $\rvx \in D$ in the feature space of $g(\cdot)$ based on their cosine similarity.
We denote the set of NN for $\rvx$ by $S_{\rvx}$. During training, TEMI randomly samples $\rvx$ from $D$ and $\rvx'$ from $S_{\rvx}$ to generate image pairs with similar (visual) features.
A self-distillation framework is introduced to learn the cluster assignments with a teacher and student head $h_t(\cdot)$ and $h_s(\cdot)$ that share the same architecture (i.e.\ 3-layer MLP) but differ w.r.t.\ their parameters.
The features of the image pair are fed to the student and teacher heads $h^{i}_s(\rvz)$, $h^{i}_t(\rvz')$, where $i \in \{ 1, \dots, H \}$ is the head index and $\rvz=g(\rvx)$, $\rvz'=g(\rvx')$.
The outputs of the heads are converted to probabilities $\pstud(c|\rvx)$ and $\pteach(c|\rvx')$ using a softmax function. The TEMI objective,  
\begin{equation}
    \begin{split}
    \temiloss^i(\rvx,\rvx') := -
     \frac{1}{H} \sum_{j=1}^H \sum_{c'=1}^C \pteach^j(c'|\rvx)\pteach^j(c'|\rvx') \\
     \log \sum_{c=1}^C \frac{
    \left( \pstud^i(c|\rvx) \pteach^i(c|\rvx')\right)^{\gamma}}
    {\tilde q_t^i(c)} \quad,
    \end{split}   
\label{eq:temi-loss}
\end{equation} 
maximizes the pointwise mutual information between images $\rvx$ and $\rvx'$, using the clustering index $c$ as information bottleneck \cite{ji2019invariant_cluster}.
Here, $\tilde\pteach^i(c)$ is an estimate of the teacher cluster distribution $\mathbb{E}_{\rvx \sim p_\text{data}}\left[\pteach^i(c|\rvx)\right]$, which can be computed by an exponential moving average over mini-batches $B$ defined as $\tilde \pteach^i(c) \leftarrow \lambda \,\tilde \pteach(c) + (1 - \lambda) \frac{1}{\abs{B}}\sum_{\rvx \in B} \pteach^i(c|\rvx) $. 

The hyperparameter $\gamma \in (0.5, 1]$ avoids the collapse of all sample pairs into a single cluster, and $\lambda \in (0,1)$ is a momentum hyperparameter. The loss function \cref{eq:temi-loss} is further symmetrized and averaged over heads to obtain the final training loss. 
After training, the cluster assignments $c^*(\rvx)=\argmax_c{\pteach(c|\rvx)}$ 
and the empirical cluster distribution $q(c)$ on the training set is computed from the TEMI head with the lowest loss. We use $c^*(\rvx)$ as a condition of the generative model during training and sample from $q(c)$ to generate images. Adaloglou et al. \cite{temi} show that the value $\gamma=0.6$ enforces a close to uniform cluster utilization and achieves state-of-the-art clustering accuracy for $C_{GT}$\cite{temi}.

\subsection{Cluster-conditional EDM (C-EDM)}
We consider $k$-means and TEMI clusters using off-the-shelf feature extractors with $C_V$ clusters to compute cluster assignments. We experimentally observed that $C_V>C_{GT}$, even though this is not a requirement of our approach. We highlight that clustering using $C>C_{GT}$ while enforcing a uniform cluster utilization (such as TEMI with $\gamma=0.6$) has not been previously explored \cite{temi,pcl,propos,scan} as it reduces the cluster alignment with the GT labels. We denote the cluster-conditioned EDM model as \emph{C-EDM}.

A direct estimation of $C_{V}$ is hard to obtain for unlabeled datasets. In principle, $C_{V}$ can be found using a hyperparameter search with FID as an evaluation metric, which is computationally expensive. In contrast to existing approaches that perform a restricted hyperparameter search around $C_{GT}$, we propose a new metric that allows us to derive an upper bound for $C_{V}$ (\cref{sec:visual-groups}), which requires no prior knowledge about the dataset. 

\begin{table*}
\centering
\begin{tabular}{l cccc cc c }
\toprule
 & \multicolumn{2}{c}{{CIFAR10}} &  \multicolumn{2}{c}{{CIFAR100}} & \multicolumn{1}{c}{{FFHQ-64}} & \multicolumn{1}{c}{{FFHQ-128}} \\
\cmidrule(lr){2-3} \cmidrule(lr){4-5} \cmidrule(lr){6-6} \cmidrule(lr){7-7}
{Generative methods} & Unlabelled & \textcolor{gray}{GT} & Unlabelled & \textcolor{gray}{GT}  &  Unlabelled & Unlabelled \\
\midrule
DDPM \cite{ho2020ddpm,hu2023self} & $3.17$ & \textcolor{gray}{-} & $10.7$ & \textcolor{gray}{$9.7$} & - & -  \\
PFGM++ \cite{xu2023pfgm} & - & \textcolor{gray}{$1.74$} & - & \textcolor{gray}{-} & $2.43$ & - \\
EDM \cite{edm} & $1.97$ & \textcolor{gray}{$1.79$} & - & \textcolor{gray}{-} & $2.39$ & - \\
EDM (our reproduction) & $2.07$ & \textcolor{gray}{$1.81$} & $3.41$ & \textcolor{gray}{$2.21$} & $2.53$ & 5.93  \\
\hline
\multicolumn{5}{l}{{\textbf{Self-conditional methods}}}  \\
IDDPM+$k$-means ($k$=10) \cite{bao2022why_cond} & 2.23  & \textcolor{gray}{-}  & -  & \textcolor{gray}{-} & - & -  \\
DDPM+$k$-means$^\dagger$ ($k$=400) \cite{hu2023self} & -  & \textcolor{gray}{-}  & $9.6$  & \textcolor{gray}{-} & - & -  \\
C-EDM+$k$-means$^\dagger$ (Ours) & 1.69 & \textcolor{gray}{-} & 2.21  & \textcolor{gray}{-} &  \textbf{1.99} & - \\
C-EDM+TEMI$^\dagger$ (Ours) & \textbf{1.67} & \textcolor{gray}{-} & \textbf{2.17}  & \textcolor{gray}{-} &  $2.09$ & \textbf{4.40} \\
\hline
\textcolor{gray}{EDM+oracle features$^\dagger$ (Ours)} & \textcolor{gray}{2.21} & - & \textcolor{gray}{2.25} & - & \textcolor{gray}{1.77} & - \\
\bottomrule
\end{tabular}
\caption{{State-of-the-art generative model comparison: FID ($\downarrow$) for various GT labeled and unlabelled benchmarks.} We use $C_V=$100, 200, 400 clusters for CIFAR10, CIFAR100, and FFHQ-64, respectively. Self-conditional methods with $^\dagger$ use the pre-trained DINO ViT-B feature extractor. Ground truth and oracle feature conditioning results are marked in \textcolor{gray}{gray color} as they are \textbf{not} a fair comparison.} 
\label{tab:sota_diffusion_models.tex}
\end{table*}

\subsection{Estimating the upper cluster bound} \label{sec:visual-groups} 
We aim to find an upper cluster bound $C_{max}$ using TEMI ($\gamma=0.6$), such that $C_{V}<C_{max}$. During this computation, we neither use the generative model nor any additional information. {We highlight that previous works \cite{cvpr2020gans_clusters, bao2022why_cond,hu2023self} have overlooked this design choice and iterate within a small range around $C_{GT}$}. We denote the number of utilized clusters (one training sample is assigned to it) as $C^{u}$ after TEMI clustering. The TEMI cluster utilization ratio is defined as $r_{C}:=\frac{C^{u}}{C} \leq 1$. Importantly, there is no guarantee that the full cluster spectrum will be utilized. Unlike $k$-means that always has $r_{C}=1$, we observe that as the number of clusters $C$ increases, the TEMI cluster utilization ratio $r_C$ typically decreases, which provides dataset-specific information. Intuitively, TEMI clustering with $\gamma=0.6$ is enforcing $r_{C}\rightarrow 1$, and the observation of $r_{C}<1$ is an indication that the maximum number of TEMI clusters is reached for $\gamma=0.6$.

The search for $C_{max}$ involves doubling the number of clusters $C$ at each iteration until the $r_C$ falls below a threshold, $r_C \leq \alpha$, followed by a more fine-grained grid search. $C_{max}$ is then defined as the highest $C$ for which $r_C > \alpha$. This results in a worst-case time complexity of $O(\log_{2}(\frac{C_{max}}{C_{\text{start}}}))$. This is conceptually similar to the elbow method \cite{thorndike1953belongs_clusterlit, ketchen1996application_clusterlit}. In this sense, the proposed heuristic has the same limitations as the elbow method. Yet, it is currently the only method that provides a practitioner with a starting point for an unlabelled dataset. After detecting $C_{max}$, we perform a bounded grid search to find $C_{V} \in [2, C_{max})$ using the generative model. To obtain a simple cross-dataset estimate, we empirically find a cutoff threshold $r_C\leq 0.96$ to work well across three datasets.

\section{Experimental evaluation}
\subsection{Datasets, models, and metrics} \label{sec:dataset_description}
Following prior works \cite{hu2023self, temi, tsp}, we use DINO ViT-B pre-trained on ImageNet \cite{deng2009imagenet} for image clustering. We report the Fréchet inception distance (FID) \cite{heusel2017fid} to quantify the image generation performance as it simultaneously captures visual fidelity and diversity. To facilitate future comparisons, we also compute the Fréchet DINOv2 distance (FDD) \cite{stein2023exposing} that replaces the features of InceptionNet-v3 \cite{szegedy2015inception,szegedy2016inceptionv3} with DINOv2 ViT-L \cite{oquab2023dinov2}. FID and FDD are averaged over three independently generated sample sets of 50K images each. To measure the cluster alignment w.r.t. GT labels, we use the adjusted normalized mutual information (ANMI) as in \cite{propos}.

We follow the default hyperparameter setup for EDM \cite{edm} and TEMI \cite{temi}. We denote the number of samples (in millions) seen during training as $M_{img}$. We use $M_{img}=200$ when comparing C-EDM with other baselines and state-of-the-art methods (\cref{tab:sota_diffusion_models.tex}), and $M_{img}=100$ when comparing across different number of clusters. We train the TEMI clustering heads for 200 epochs per dataset. All the experiments were conducted on 4 NVIDIA A100 GPUs with 40GB VRAM each. On this hardware, TEMI clustering was more than $50\times$ faster than training EDM on FFHQ-64, which requires more than $2$ days for 200 $M_{img}$ with a batch size of $512$. Additional implementation details and hyperparameters can be found in the supplementary material. We verify our approach across CIFAR10, CIFAR100 \cite{cifar} and FFHQ. CIFAR10 and CIFAR100 have 50K samples and $32^2$ resolution images, while we use the 64x64 and 128x128 versions of FFHQ (FFHQ-64, FFHQ-128) consisting of 70K samples. Finally, diffusion sampling methods are not included, as they can be incorporated into any diffusion model \cite{kim2022refining,sadat2023cads,ho2022cfg}.

\begin{figure*}
\begin{center}
  \includegraphics[width=1.85\columnwidth]{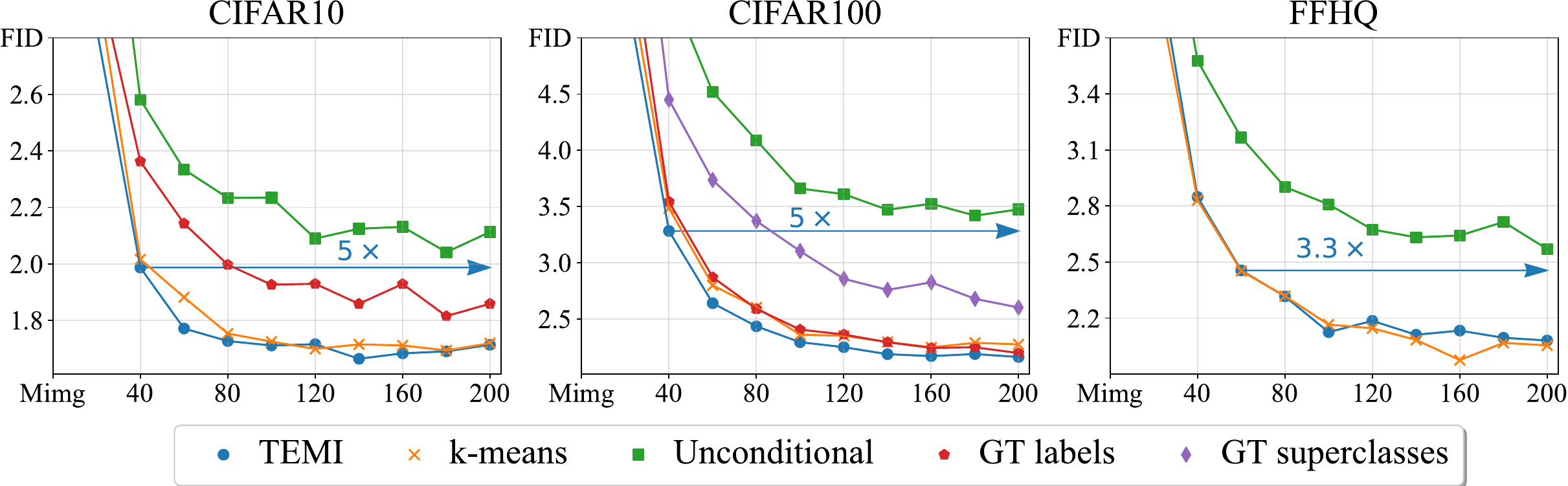}
    \caption{FID (\textit{y-axis}) versus seen samples during training in millions (\textit{x-axis}). TEMI and k-means clusters are computed using the representations of DINO ViT-B \cite{dino}. We used $C_V=100,200,400$ for CIFAR10, CIFAR100 and FFHQ-64 respectively. The training sample efficiency compared to the unconditional baseline is indicated by the arrow. Best viewed in color.}
    \label{fig:sample_eff}
\end{center}
\end{figure*}

\subsection{State-of-the-art comparison for image synthesis} \label{sec:sota}

\begin{figure*}
\begin{center}
  \includegraphics[width=1.85\columnwidth]{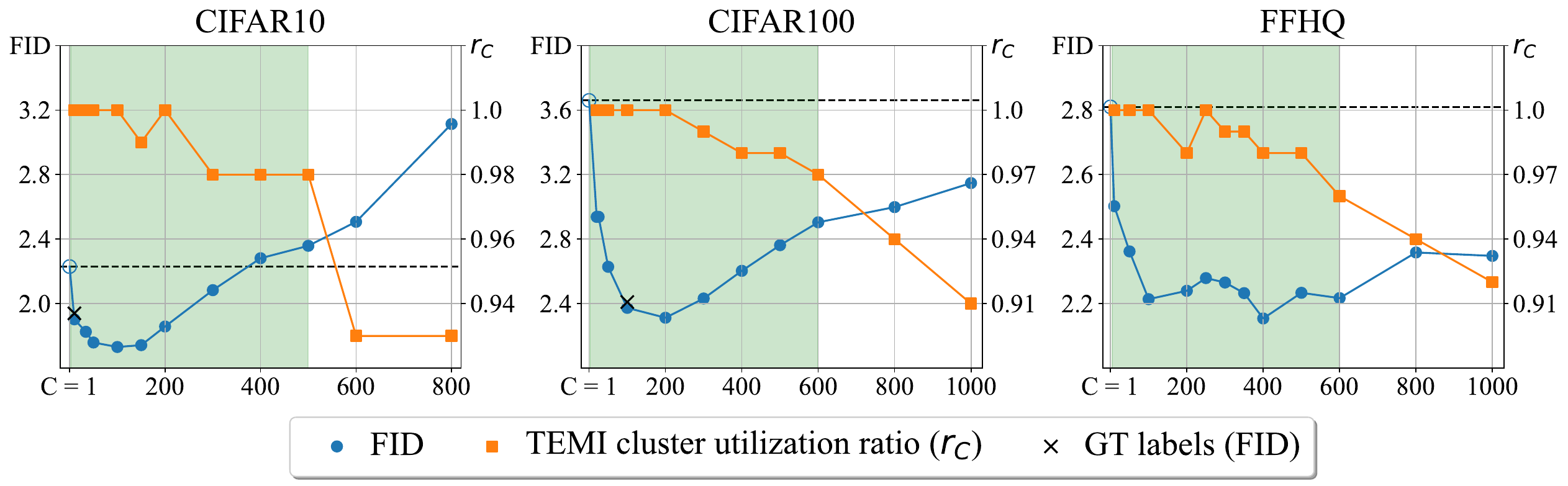}\caption{FID (\textit{left y-axis}) and TEMI cluster utilization ratio $r_C$ (\textit{right y-axis}) across different numbers of clusters $C$ (\textit{x-axis}) using C-EDM, evaluated at $M_{img}=100$. The green area indicates the discovered cluster range $[2,C_{max})$ for $r_C\leq \alpha=0.96$.}\label{fig:utlization-ration}
\end{center}  
\end{figure*}

\textbf{Comparison with state-of-the-art unconditional generative models.} {With respect to FID, we found $C_V=100,200,400$ to be close to optimal cluster granularities for CIFAR10, CIFAR100, and FFHQ-64, respectively. We use these values to report FIDs compared to GT conditional and unconditional generative models in \cref{tab:sota_diffusion_models.tex}. Compared to unconditional methods, C-EDM achieves an average relative FID improvement of 24.4\% and 24.9\% using TEMI and k-means clusters, respectively.} More importantly, previous cluster-conditional approaches did not achieve near-state-of-the-art FIDs because they: a) adopted non-competitive diffusion baselines such as DDPM \cite{ho2020ddpm}, b) did not consider pre-trained feature extractors for clustering \cite{bao2022why_cond}, c) did not use the optimal cluster granularity \cite{hu2023self}. For instance, Hu et al. \cite{hu2023self} used $400$ clusters on CIFAR100 using DINO ViT-B for clustering, while \cref{fig:utlization-ration} shows that $200$ clusters lead to a superior FID using C-EDM (11.2\% relative improvement). 

\textbf{Comparison with state-of-the-art GT conditional generative models.} {Intriguingly, using C-EDM with $C_V$ clusters, we report small improvements compared to GT label conditioning on CIFAR10 and CIFAR100 in \cref{tab:sota_diffusion_models.tex}. For instance, 4.77\% mean relative improvement in FID using TEMI. Even though cluster-conditioning is primarily designed for unconditional generation, such as FFHQ, we demonstrate that it can match or outperform GT labels, which showcases the effectiveness of C-EDM.}

\noindent\textbf{Sample efficiency.} In addition to the reported gains in FID, we study the sample efficiency of C-EDM during training across datasets in \cref{fig:sample_eff}. On CIFAR10 and CIFAR100, the training sample efficiency for $C=C_V$ compared to the unconditional model ($C=1$) peaks at $5\times$, where C-EDM with $M_{img}=40$ outperforms the unconditional model at $M_{img}=200$. On FFHQ-64, which is not a classification dataset, we report a sample efficiency of $3.3\times$. TEMI and $k$-means demonstrate identical sample efficiency compared to the unconditional model. {Intuitivly, a more informative conditioning signal enables learning the data distribution faster. Upon visual inspection, we could identify FFHQ clusters for $C=C_V$ with easily distinguishable visual patterns, such as groups of images with beanies, smiling faces, glasses, sunglasses, hats, and kids (see supplementary).} Finally, we find that the sample efficiency of C-EDM against GT conditional EDM heavily depends on the quality and granularity of the GT labels.

\subsection{Cluster utilization ratio and discovered upper bounds} \label{sec:cluster_util_results} We depict how the TEMI cluster utilization $r_C$ changes in tandem with FID for different numbers of clusters in \cref{fig:utlization-ration}. Crucially, this approach also applies to FFHQ, where no GT labels exist. Moreover, we can discern certain patterns across datasets: (i) $C_V$ always has a high utilization ratio, (ii) the majority of experiments with $C<C_{max}$ outperform the unconditional model at $M_{img}=100$, and (iii) $C_V>C_{GT}$ for CIFAR10 and CIFAR100, which is in line with \cite{hu2023self}. {Even though the choice of $\alpha=0.96$ is not guaranteed to be generally applicable and is based on empirical evidence, a more or less strict choice can be used based on the practitioner's computational budget.} The introduced upper bound can be computed with negligible computational overhead and without access to GT labels or training the generative model, which on large scales (ImageNet) can require up to $4$MWh per experiment \cite{karras2023edm2}.

\subsection{Investigating the connection between clustering and cluster-conditional image synthesis} \label{sec:analysis}
\noindent\textbf{FID and ANMI.} On image clustering benchmarks, TEMI ($\gamma=0.6$) outperforms $k$-means, where we respectively measure an ANMI of 60.6\% versus 59.2\% on CIFAR10 and 72.3\% versus 67.6\% on CIFAR100. However, their generative performance displays negligible differences in terms of FID for $C=C_{V}$. We emphasize that the imbalanced clusters of $k$-means are penalized when computing clustering metrics such as ANMI on balanced classification datasets, while during image synthesis, this is naturally mitigated by sampling from $q(c)$. \cref{tab:sota_diffusion_models.tex} suggests that imbalanced cluster assignments are beneficial for unlabelled datasets such as FFHQ, where we report a relative gain of 4.8\% using k-means clusters compared to TEMI.

\begin{table}
\begin{center}
\begin{tabular}{l ccc}
\toprule
 &  \multicolumn{3}{c}{{$C$}}  \\
\cmidrule(lr){2-4}
 CIFAR100 &  \multicolumn{1}{c}{{20}}   &  \multicolumn{1}{c}{{100}} &  \multicolumn{1}{c}{{200}}  \\
\midrule
Human annotation (GT) & 3.10 & 2.41  & - \\
CLIP pseudo-labels & 3.38 & 2.42 &  -  \\
$k$-means clusters & 3.09  & 2.41 & 2.36 \\
TEMI clusters & \textbf{2.93} & \textbf{2.37} &  \textbf{2.31} \\
\bottomrule
\end{tabular}
 \caption{FID for different grouping methods as a condition to EDM
($M_{img}=100$). We use the 20 GT superclasses and 100 GT labels of CIFAR100. CLIP pseudo-labels are computed using zero-shot classification pseudo-labels based on the GT label names.}\label{tab:grouping_cifar100_annotations}
 \end{center}

\end{table}

\noindent\textbf{C-EDM matches EDM when $C=C_{GT}$.} In \cref{fig:utlization-ration}, we observe that the GT label conditioning closely follows the FID of TEMI clusters for $C=C_{GT}$. In \cref{tab:grouping_cifar100_annotations}, we leverage the two annotation levels of CIFAR100 \cite{cifar}, specifically the 20 GT superclasses and the 100 GT labels to benchmark how different grouping methods perform in image synthesis. 
Apart from image clustering, we create text-based pseudo-labels with CLIP (OpenCLIP ViT-G/14 \cite{openclip}) similar to \cite{ming2022delving}. Then, CLIP pseudo-labels are derived from zero-shot classification \cite{radford2021clip} using the $\argmax$ of the text-image similarity after softmax \cite{adaloglou2023adapting}. For reference, we provide the FID with the number of visual groups $C_V=200$ for $k$-means and TEMI clusters. Interestingly, all methods attain a similar FID for $C_{GT}=100$, while TEMI achieves the best FID only for $C=20$. This suggests that generative performance is mostly invariant to the grouping method, including human annotation, given a feature extractor that captures general visual concepts. Next, we investigate the impact of the pre-trained feature extractor.

\begin{figure}     
\centering
    \includegraphics[width=0.9\columnwidth]{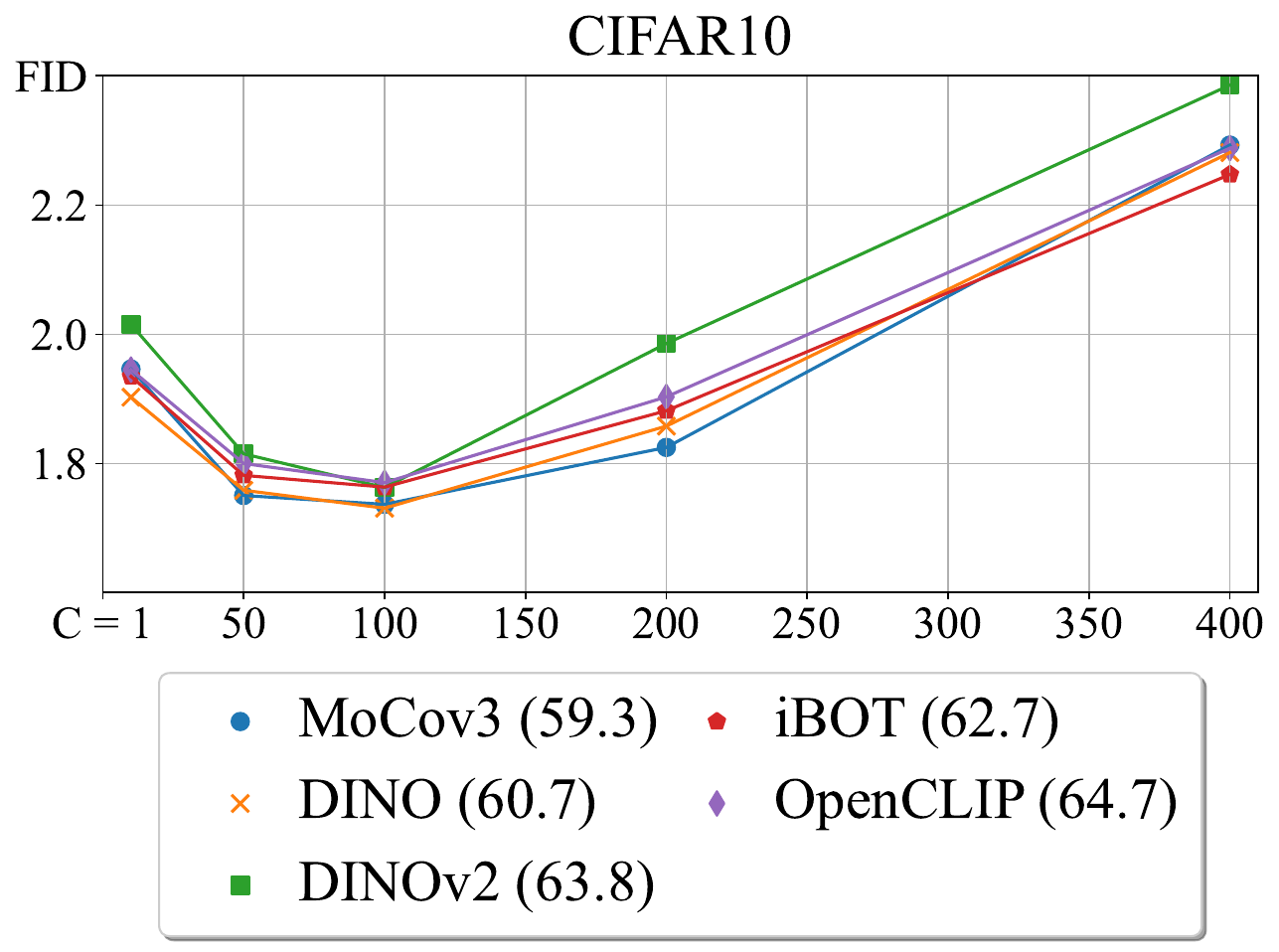}
    \captionof{figure}{FID (\textit{y-axis}) across different numbers of clusters $C$ (\textit{x-axis}) using C-EDM with TEMI with different feature extractors. The ANMI is shown in parentheses for $C_V$=100.} \label{fig:backbones-cifar10-cluster-range}
\end{figure}

\begin{table*} 
\centering
\begin{tabular}{l cccc  cccc  }
\toprule
 & \multicolumn{4}{c}{{CIFAR10 ($C=100$)}}  & \multicolumn{4}{c}{{CIFAR100 ($C=200$) }}  \\
\cmidrule(lr){2-5} \cmidrule(lr){6-9}
{C-EDM+TEMI} & ACC$\uparrow$ & ANMI$\uparrow$ & FID$\downarrow$ & FDD$\downarrow$ &  ACC$\uparrow$ & ANMI$\uparrow$ & FID$\downarrow$ & FDD$\downarrow$  \\
\midrule
DINO ImageNet weights  & 96.4 & 60.6 & \textbf{1.73}  & 153.5 & 82.3  & 72.3 & \textbf{2.31}  & \textbf{250.7} \\
DINO adapted weights & \textbf{98.5} & \textbf{63.41} &  1.74 & \textbf{152.8} &  \textbf{87.8} & \textbf{78.2} & 2.39   & 250.8 \\
\bottomrule
\end{tabular}
\caption{Effect of self-supervised fine-tuning (adapted weights) on the pre-trained DINO ViT-B. ACC stands for the 20-NN classification accuracy using the DINO features. We measure FID and FDD for both models at $M_{img}=100$.}
\label{tab:ssl_ft_tuning.tex}
\end{table*}

\noindent\textbf{The number of visual groups is insensitive to the chosen feature extractor.} The broader adoption of the discovered $C_V$ for each dataset requires our analysis to be insensitive to the chosen feature extractor. In \cref{fig:backbones-cifar10-cluster-range}, we show that the pre-trained models considered in this work achieve similar FIDs across different numbers of clusters even though the ANMI varies up to 5.4\% on CIFAR10. Feature extractors differ in terms of the number of parameters (86M up to 1.8 billion), pre-training dataset size (1.2M up to 2B), and objective (MoCO\cite{mocov3}, DINO\cite{dino}, iBOT\cite{zhou2021ibot}, CLIP \cite{openclip,radford2021clip}). {Our finding is consistent with concurrent work on the scale of ImageNet \cite{meta2023gen_representations} that shows that linear probing accuracy is not indicative of generative performance for feature-based conditioning.} The fact that differences in discriminative performance do not translate to improvements in FID is in contrast with Hu et al. \cite{hu2023self}.

\noindent\textbf{Dataset-specific features do not improve generative performance.} To realize dataset-dependent adaptation of the feature extractor, we perform self-supervised fine-tuning on CIFAR10 and CIFAR100, starting from the ImageNet weights in \cref{tab:ssl_ft_tuning.tex}. We train all layers for 15 epochs similar to \cite{ssl-ft} using the DINO framework, resulting in ``\emph{DINO adapted weights}''. We find large improvements in classification accuracy and ANMI, which is in line with the fact that the learned features are more dataset-specific \cite{rafiee2022self,ssl-ft,wang2023clipn}. However, the gains in classification and clustering do not translate into conditional image generation, which suggests that image synthesis requires features that capture general visual concepts. 

\begin{figure*}
\begin{center}
  \includegraphics[width=1.9\columnwidth]{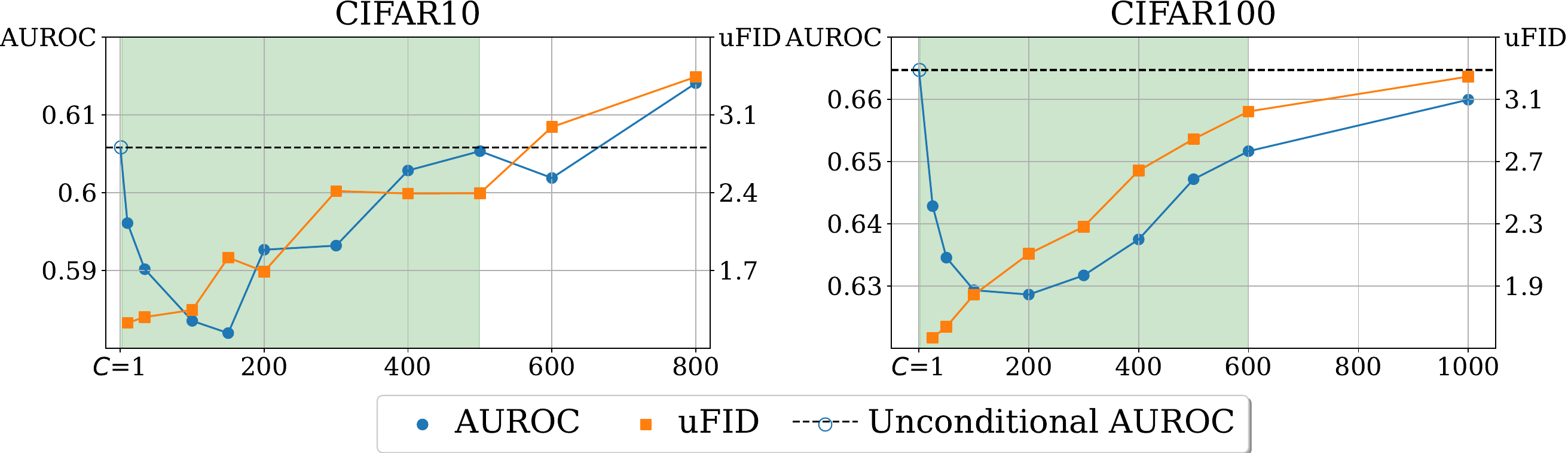}
    \caption{Top 1-NN cosine similarity AUROC (\textit{left y-axis}) and Frechet distance between the C-EDM and unconditional samples (uFID) for different cluster sizes $C$ (\textit{x-axis}). For the computation of AUROC, we use the official test splits.} 
    \label{fig:auroc_cos-sim}
\end{center}
\end{figure*}

\section{Discussion}
\textbf{Highly fine-grained clusters lead to out-of-distribution samples.} To quantify the degree of out-of-distribution for the generated C-EDM samples, we measure the AUROC using the top-1 NN cosine similarity (1-NN) using DINO ViT-B \cite{sun2022out} and compare with the test split of CIFAR10 and CIFAR100 (\cref{fig:auroc_cos-sim}). We note that 1-NN is independent of the data distribution of the generated samples and thus does not account for diversity. The highest $C$ considered for CIFAR10 and CIFAR100 produces the highest AUROC, suggesting that the features of the generated samples are pushed away from the training data. We hypothesize that this behavior originates from highly specialized clusters that cannot always be generated from the initial noise, resulting in visual artifacts. In parallel, we measure how similar the C-EDM samples are compared to the unconditional ones using FID. We call this metric \emph{unconditional FID (uFID)}. We observe that uFID increases as $C$ increases, suggesting that the generated samples for larger $C$ are farther away in feature space than the unconditional samples.

\noindent\textbf{The clusters' granularity level determines conditional generative performance.} The quality of feature representations is typically determined by linear separability w.r.t. GT labels \cite{park2023what_ssl_vit}. However, our experimental analysis shows that in image synthesis, the quality of image-level conditioning primarily depends on the granularity of the cluster assignments (\cref{tab:grouping_cifar100_annotations} and \cref{fig:utlization-ration}). Overall, we find no significant connection between the task of feature-based clustering and cluster-conditional image generation (\cref{fig:backbones-cifar10-cluster-range} and \cref{tab:ssl_ft_tuning.tex}). The pre-trained model and the categorization method do not severely affect generative performance for the considered datasets.

\begin{figure*}
\begin{center}
  \includegraphics[width=1.85\columnwidth]{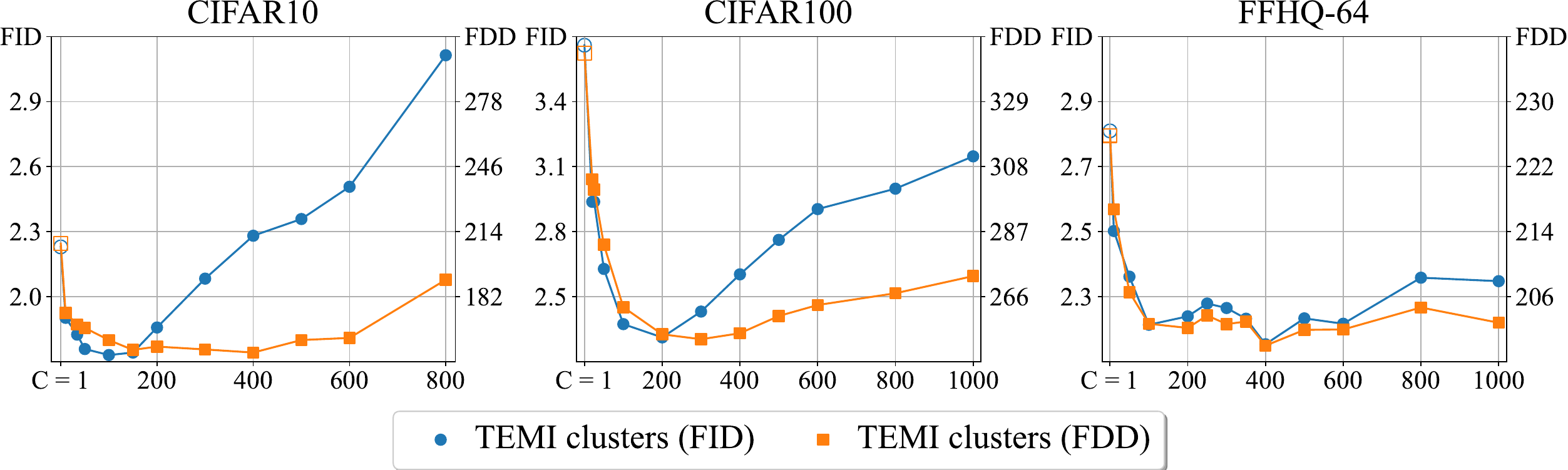}
    \caption{FID (\textit{left y-axis}) and FDD (\textit{right y-axis}) versus number of clusters (\textit{x-axis})  using C-EDM at $M_{img}=100$. The non-filled markers indicate the unconditional EDM.} \label{fig:fid_fdd_clusters_new}
\end{center}
\end{figure*}

\noindent\textbf{Generative metric and optimal number of clusters.} Here, we investigate the impact of the choice of generative metric by comparing FDD and FID. We highlight that the only difference is that FID and FDD use InceptionV3 and DINOv2 features, respectively. The cross-dataset evaluation in \cref{fig:fid_fdd_clusters_new} shows that FID and FDD do not always agree with respect to the number of visual groups. The largest disagreement is observed on CIFAR10. Still, the discovered upper bound $C_{max}$ always includes the number of visual groups. Moreover, we evaluate both metrics across training iterations and found that while FID fluctuates after $M_{img}=120$, FDD decreases monotonically (supplementary material). Finally, we measured additional feature-based metrics such as precision, recall, diversity, coverage, MSS, and inception score without success. Similar to \cite{chong2020effectively, kynkaanniemi2022role, parmar2022aliased,jayasumana2023rethinking,stein2023exposing}, we argue that new generative metrics must be developed.

\begin{figure}
    \centering
    \includegraphics[width=0.95\columnwidth]{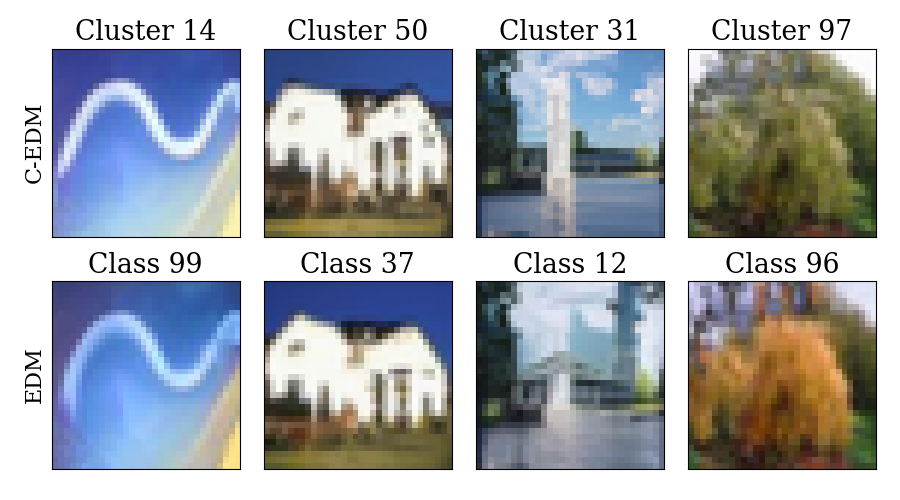}
    \caption{Comparing C-EDM samples generated with learned cluster assignments (top row, C=100) to EDM samples generated with GT labels, on CIFAR100. %
    The clusters are mapped to GT classes using the Hungarian algorithm.}
    \label{fig:visual_comp}
\end{figure}

\noindent\textbf{Visual comparison of C-EDM to EDM.} In \cref{fig:visual_comp}, we map clusters to their respective CIFAR100 classes and produce samples with C-EDM and GT conditional EDM using the same noise. Samples exhibit high visual similarity, corroborating with \cref{tab:grouping_cifar100_annotations} and \cref{fig:sample_eff}. More visualizations are provided in the supplementary material.

\begin{figure}
    \centering
    \begin{subfigure}{1\columnwidth}
        \includegraphics[width=\textwidth]{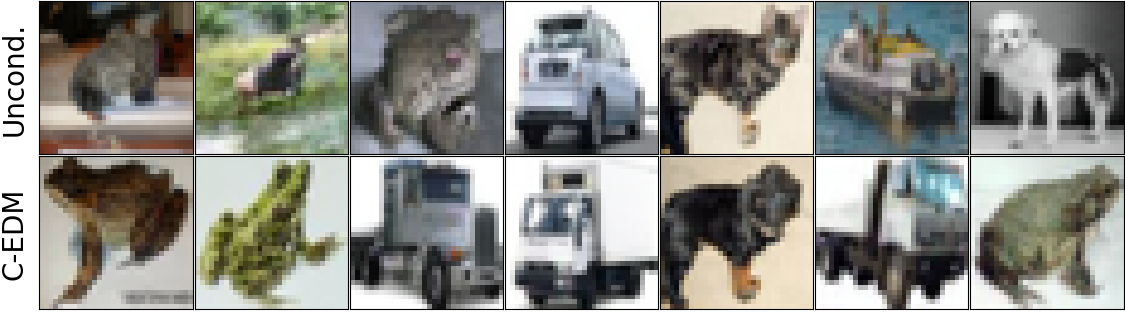}
        \caption*{(a) High-confidence C-EDM samples}
    \end{subfigure}%
    \\
    \begin{subfigure}{1\columnwidth}
        \includegraphics[width=\textwidth]{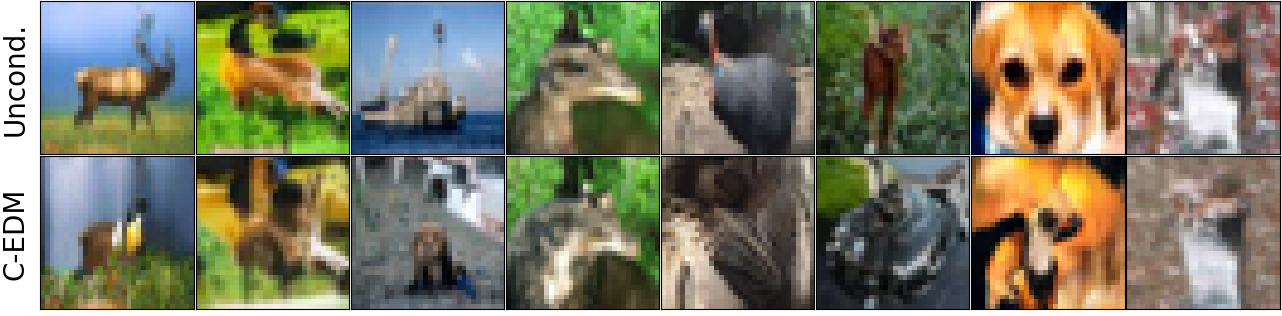}
        \caption*{(b) Low-confidence C-EDM samples}
    \end{subfigure}%
    \caption{Generated high (\textit{a}) and low (\textit{b}) confidence CIFAR10 samples. The top row depicts the unconditional (\textit{Uncond.}) samples, while the bottom row shows the generated samples using \textit{C-EDM} with TEMI ($C=100$). Images on the same column are produced with the same initial noise.}
    \label{fig:low_confidence_samples_all_data}
\end{figure}

\noindent\textbf{Measuring the confidence of the generated samples using TEMI.} 
In \cref{fig:low_confidence_samples_all_data}, we show the generated examples with the lowest and highest maximum softmax probability \cite{hendrycks2017msp} of the TEMI head as a measure of confidence. For comparison, we show unconditional samples generated using the same initial noise in the denoising process. Visual inspection shows that low-confidence C-EDM samples do not have coherent semantics compared to the unconditional ones, leading to inferior image quality. {We hypothesize that the sampled condition for the low-confidence samples is in conflict with the existing patterns in the initial noise. Increasing the number of conditions likely leads to worse image-condition alignment \cite{saharia2022photorealistic,ho2022cfg, diffusion-self-attention2023iccv}. We argue that internal guidance methods could be employed to increase the image quality in such cases, which is left for future work.}

By contrast, highly confident C-EDM samples show more clearly defined semantics than unconditional ones. When the low frequencies, such as the object's shape, remain intact, cluster conditioning aids in refining local pixel patterns. Confident C-EDM samples consist of simple pixel patterns that are easy to generate, such as a white background. Confidence can also be leveraged in future works in rejection sampling schemes \cite{casella2004generalized}. More samples are provided in the supplementary material.

\section{Conclusion}
In this paper, a systematic empirical study was conducted focusing on conditioning diffusion models with cluster assignments. It was demonstrated that cluster conditioning achieves state-of-the-art FID on three generative benchmarks while attaining strong sample efficiency. To reduce the search space for estimating the visual groups of the dataset, a novel method that computes an upper cluster bound based solely on clustering was proposed. Finally, our experimental study indicates that generative performance using cluster assignments depends primarily on the granularity of the assignments.

{\small
\bibliographystyle{ieee_fullname}
\bibliography{main}
}

\newpage
\clearpage
\appendix

\textbf{Code.} Code is available at \url{https://github.com/HHU-MMBS/cedm-official-wavc2025}.

\section{Visualizations} \label{app:viz}

In \cref{fig:edm_vs_cedm_comp}, we map the TEMI clusters to classes using the Hungarian mapping \cite{kuhn1955hungarian} on CIFAR100. For the mapping to be one-to-one, we set $C=100$. We then generate samples using the same initial noise with both C-EDM and GT-conditioned EDM and visualize the first 20 clusters on CIFAR100. Given the same initial noise and the cluster that is mapped to its respective GT class, we observe a lot of visual similarities in the images, even though the two models (C-EDM and EDM) have different weights and have been trained with different types of conditioning. 

\begin{figure*}
    \centering  \includegraphics[width=2\columnwidth]{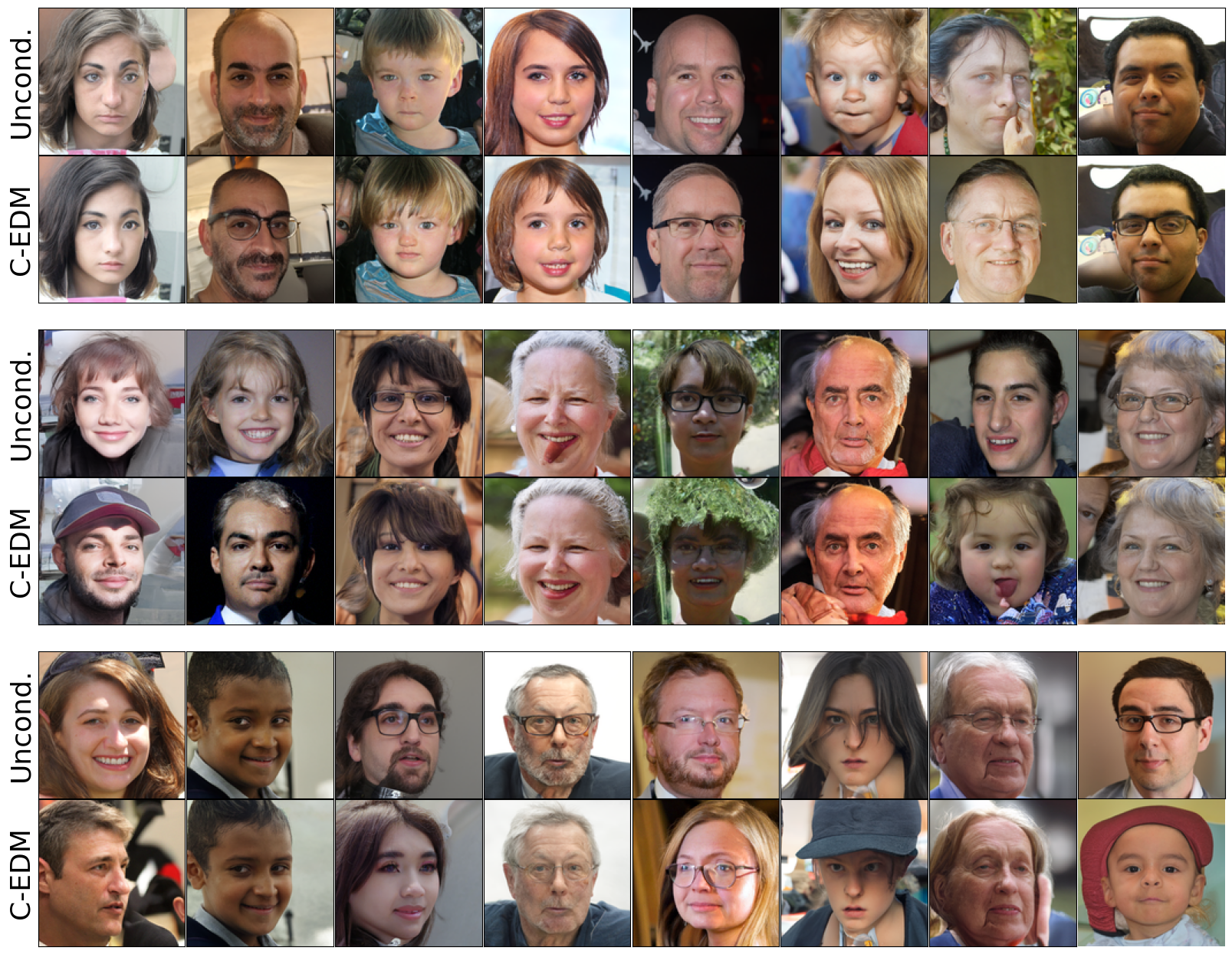}
  \caption{Visual comparison between C-EDM (C=400) and unconditional EDM (Uncond.) at $M_{img}=100M$ on FFHQ at 128x128.}
    \label{fig:ffhq128}
\end{figure*}

\begin{figure*}
\begin{center}
  \includegraphics[width=2\columnwidth]{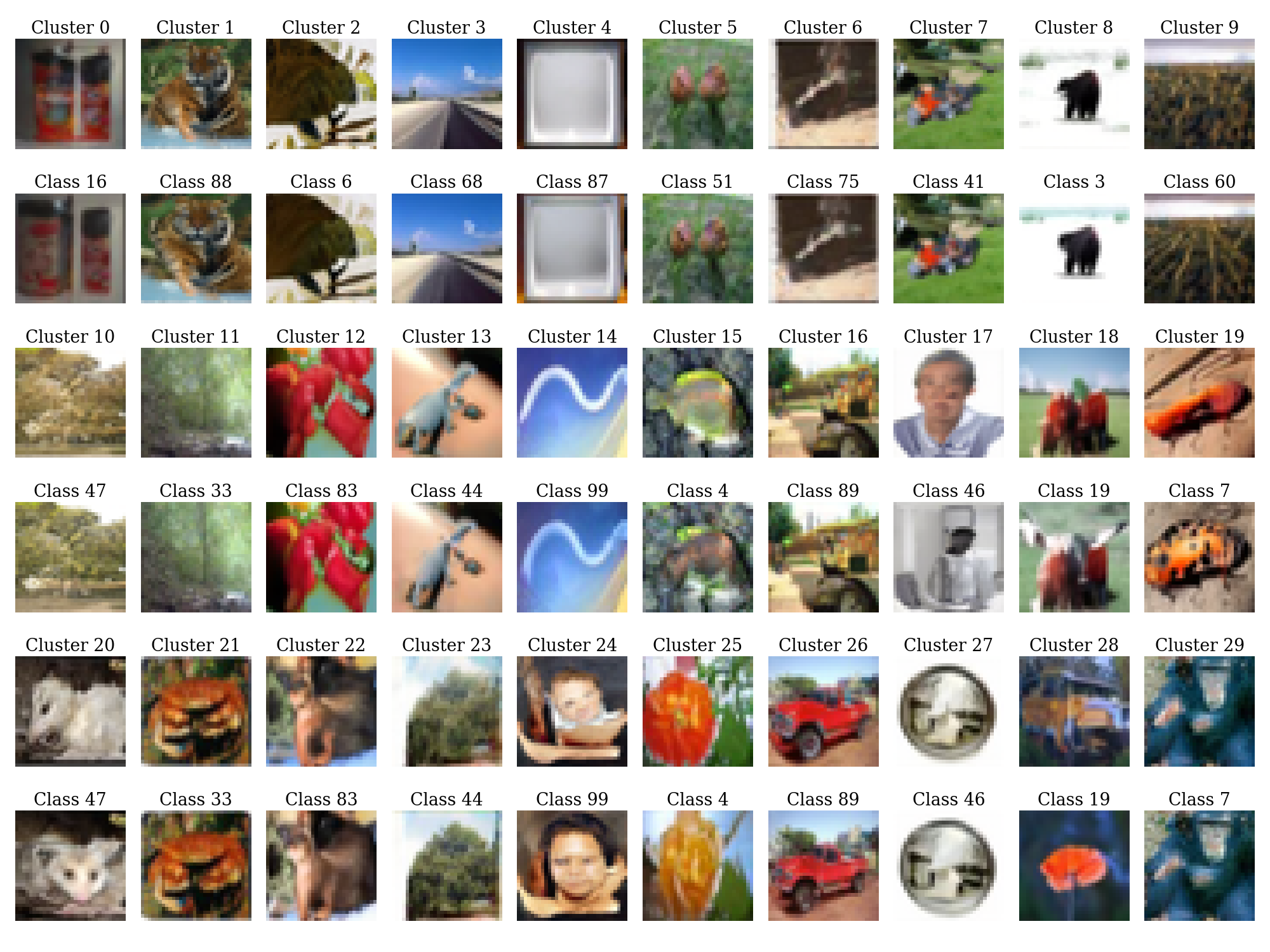}\caption{Visualizing generated images from CIFAR100 using C-EDM (even rows) and ground truth conditional EDM (odd rows) using the same initial noise and deterministic noise sampling. We map the C=100 CIFAR100 cluster to the respective ground truth class as computed via the Hungarian one-to-one mapping.}
    \label{fig:edm_vs_cedm_comp}
\end{center}
\end{figure*}

\begin{figure*}
\begin{center}
  \includegraphics[width=2\columnwidth]{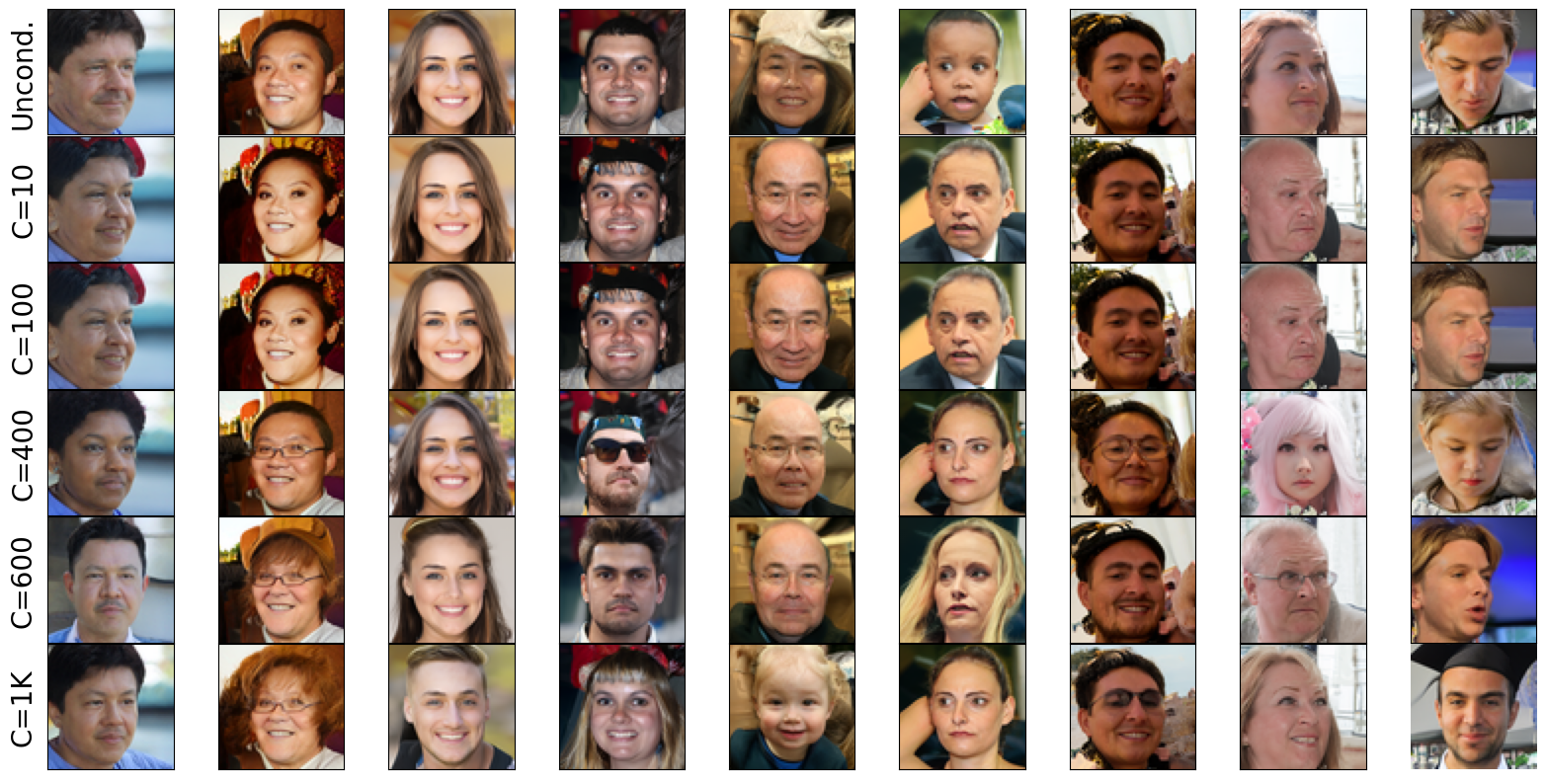}\caption{{Visualizing \textit{generated} images from FFHQ-64 using CEDM for different number of clusters $C$ with the same random noise.} We use deterministic noise sampling. Each noise gets a condition sampled from $p(c)$ for each individual clusters.}
    \label{fig:temi_clusters_vis_ffhq}
\end{center}
\end{figure*}

\begin{figure*}
\begin{center}
  \includegraphics[width=2\columnwidth]{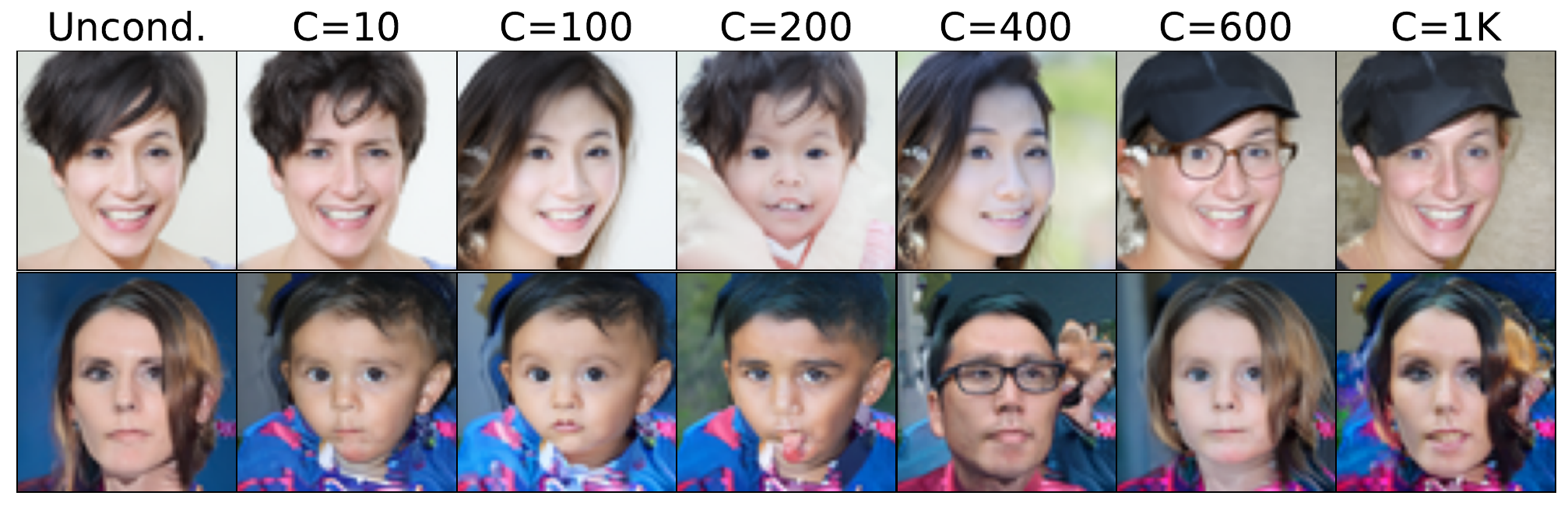}
  \caption{Generated FFHQ-64 samples using \textit{C-EDM} and TEMI clusters with different granularity levels $C$ as well as unconditional EDM (\textit{Uncond.}, first column). All samples in a row use the same initial noise. The cluster assignment is randomly sampled from $q(c)$ for each $C$.}
    \label{fig:teaser}
\end{center}
\end{figure*}

In \cref{fig:temi_clusters_vis_ffhq,fig:teaser}, we visualize C-EDM samples generated from the same initial noise on FFHQ-64 for diffusion models trained with varying cluster granularities. Each noise gets a condition sampled from $p(c)$. Similar to our quantitative analysis, the generated images from small cluster sizes are closer to the unconditional prediction. Finally, in \cref{fig:ffhq128}, we visualize cluster-conditioned and unconditional FFHQ samples at $M_{img}=100M$. 

\begin{figure*}
\begin{center}
  \includegraphics[width=2\columnwidth]{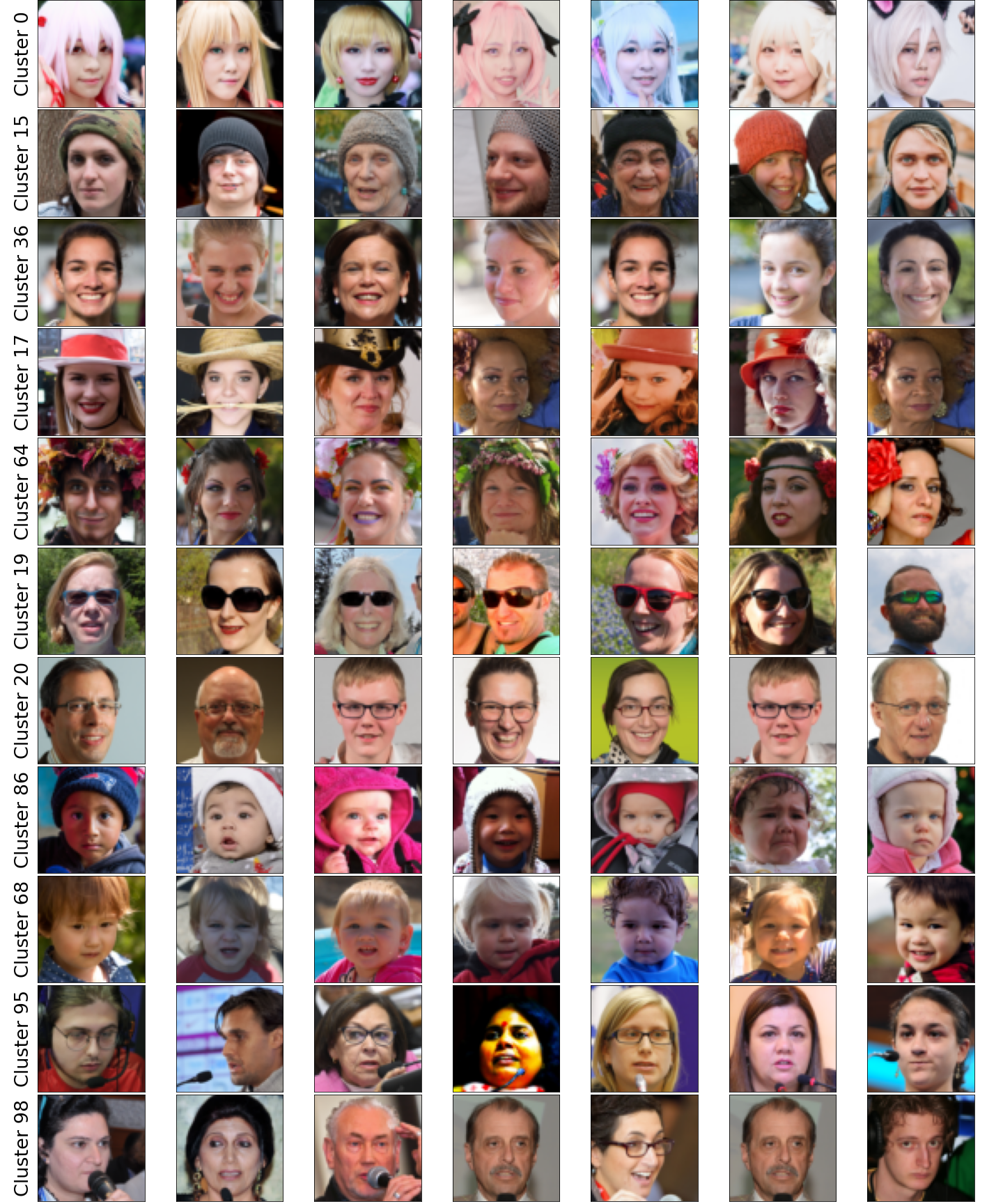}\caption{{Visualizing \textit{training} images from FFHQ that belong to the same TEMI cluster.} Images that are grouped into the same cluster are shown in the same row. We use the trained TEMI model with $C_V=400$ using the DINO backbone. Cluster assignments are picked to illustrate that images with similar visual characteristics are grouped together (i.e., beanies, smiling faces, glasses, hats, kids, etc.). Images are \textbf{randomly} sampled from each cluster.}
    \label{fig:train_samples_ffhq_temi}
\end{center}
\end{figure*}

\begin{figure*}
    \centering
    \begin{subfigure}{2\columnwidth}
        \includegraphics[width=\textwidth]{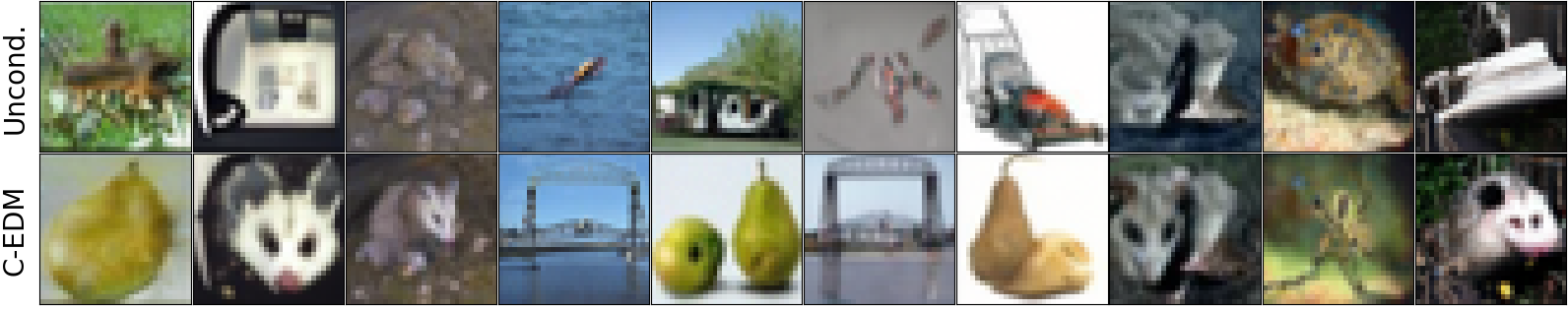}
        \caption*{(a) High-confident C-EDM samples}
    \end{subfigure}%
    \\
    \begin{subfigure}{2\columnwidth}
        \includegraphics[width=\textwidth]{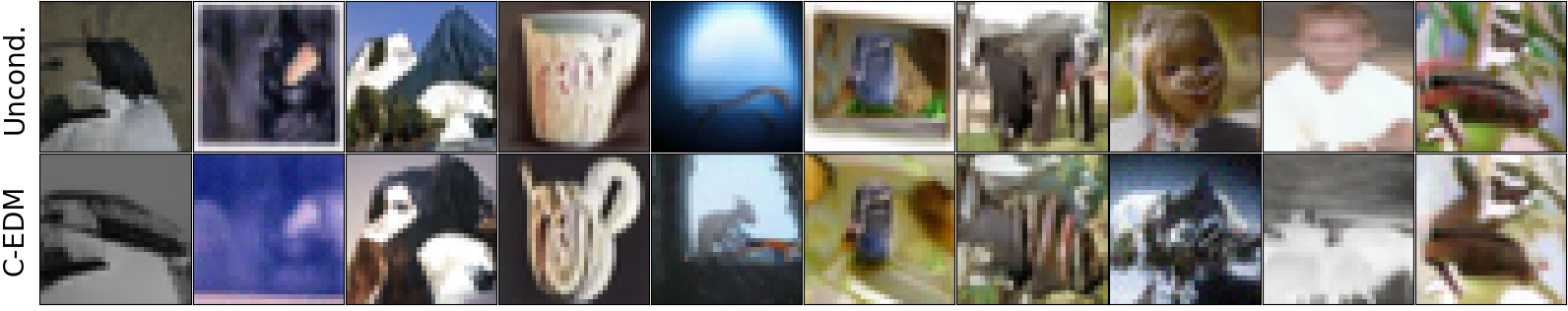}
        \caption*{(b) Low-confident C-EDM samples}
    \end{subfigure}%
    \caption{Generated low- (\textit{a}) and high-confident (\textit{b}) \textbf{CIFAR100} samples . The top row depicts the unconditional (\textit{Uncond.}) samples, while the bottom row shows the generated samples using \textit{C-EDM} with TEMI ($C=200$). Images on the same column are produced with the same initial noise. Confidence is quantified using maximum softmax probability (MSP). MSP is measured using TEMI trained on CIFAR100 without annotated data.}
    \label{fig:sup_cifar100_confidence}
\end{figure*}

In \cref{fig:train_samples_ffhq_temi}, we visualize real training FFHQ images that are grouped in the same TEMI cluster using the DINO features. We visually identify groups with shared characteristics such as sunglasses, hats, beanies, pictures of infants, and pictures of speakers talking to a microphone. Finally, in \cref{fig:sup_cifar100_confidence} we provide a more detailed visual comparison of low and high confidence samples using C-EDM on CIFAR100.

\section{Additional discussion points}

\subsection{FID and FDD across training iterations} 
\begin{figure*}
\begin{center}
  \includegraphics[width=2\columnwidth]{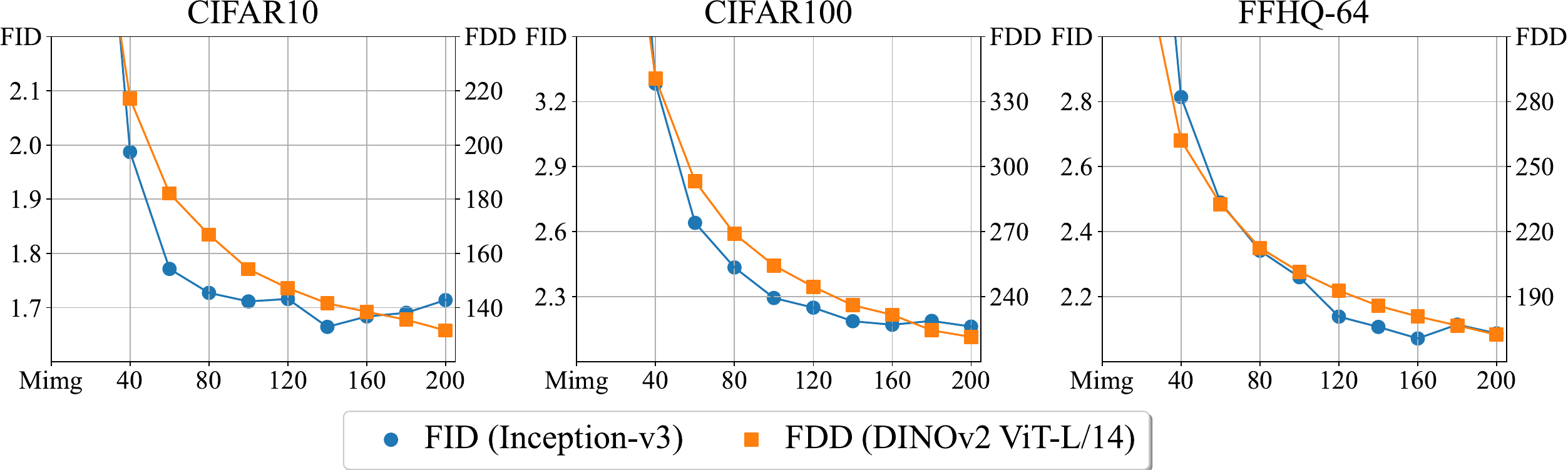}
    \caption{FID score (\textit{y-axis, left}) and FDD (\textit{y-axis, right}) during training samples seen ($M_{img}$, \textit{x-axis}). We used $C_V=100,200,400$ for CIFAR10, CIFAR100 and FFHQ-64 respectively.}
    \label{fig:gen_metrics_training}
\end{center}
\end{figure*}

In \cref{fig:gen_metrics_training}, we report FID and FDD across training using C-EDM with TEMI clusters. We notice that FID tends to saturate faster than FDD and fluctuates more between checkpoints. FDD keeps decreasing monotonically, with minimal fluctuation and always prefers the samples at $M_{img}=200$. Since both metrics compute the Frechet distance, these tendencies can only be attributed to the supervised InceptionV3 features. Even though the study of generative metrics is out of the scope of this work and a human evaluation is necessary as in \cite{stein2023exposing}, we hope that our findings w.r.t. cluster-conditioning can facilitate future works.

\subsection{Image synthesis beyond ImageNet.} ImageNet is currently the largest labeled public dataset, and a single experiment using a recent state-of-the-art diffusion model on ImageNet requires up to 4MWh at $512^2$ resolution \cite{karras2023edm2}. Based on our experiments, clusters match or outperform the human-derived labels on image generation by estimating the visual groups. Using the introduced upper bound, the search space of the visual groups is significantly reduced with minimal computational overhead, while no further hyperparameter tuning is required. Therefore, it allows future works to incorporate unlabelled data and experiment at scales beyond ImageNet while being sample efficient. Additionally, the sample efficiency compared to noisy or non-mutually exclusive labels could be investigated in future works.  

\section{Deep image clustering with TEMI}

\subsection{Intuition for $\gamma$}
In the TEMI loss function, there are two parts inside the $\log$ sum: the numerator $\left( \pstud^i(c|x) \pteach^i(c|x')\right)^\gamma$ aligns the cluster assignment of a positive pair and is maximal when each individual assignment is one-hot. On the other hand, the denominator $\tilde q_t^i(c)$ promotes a uniform cluster distribution. By dividing element-wise with the cluster probability, it is effectively up-weighing the summand corresponding to classes with low probability. In other words, when $\tilde q_t^i(c)$ is low. The hyperparameter $\gamma$ reduces the influence of the numerator, which leads to partial collapse \cite{iclr2024partial_collapse} when $\gamma=1$.

\subsection{What about the lower bound? TEMI with $\gamma=1$ experiments.}
Starting with a high overestimation of the number of clusters (e.g. $1K$ for CIFAR10), we find that TEMI clustering with $\gamma=1$ utilizes a subset of clusters, which could be used as a lower cluster bound. More precisely, we find a maximum standard deviation of $6.4$ for $C^{u}$ across datasets and feature extractors (see Supp.). Intuitively, $C^{u}$ is the minimum amount of clusters TEMI (with $\gamma=1$) uses to group all image pairs. This behavior is analogous to cluster-based self-supervised learning (using image augmentations) \cite{zhou2021ibot,dino} and has been recently coined as partial prototype collapse \cite{iclr2024partial_collapse}. Nonetheless, the lower bound is more applicable to large scales as the measured standard deviation might exclude the optimal granularity for small, highly curated datasets. Due to the above limitation, we leave this for future work. 

As depicted in \cref{tab:backbones_beta_1_cifar10}, the utilized number of clusters $C^u$ is not sensitive to the pre-determined number of clusters nor the choice of backbone for TEMI clustering when $\gamma=1$.

\begin{table*}%
\begin{center}
\caption{Number of utilised clusters $C^u$ for different number of input clusters $C$ (\textit{left}) and different backbones (\textit{right}) using TEMI with $\beta=1$ with the DINO ViT-B/16 backbone. We show the relatively small sensitivity of $C^u$ to the choice of $C$ and backbone; we report a standard deviation of a maximum value of 6.37 across different cluster sizes and 6.44 across backbones on CIFAR10.} 
\begin{minipage}{0.45\textwidth}
\begin{tabular}{lccc}
\toprule
 TEMI  &  CIFAR10 &  FFHQ &  CIFAR100  \\
  $\gamma=1$ &  $C^{u}$ &  $C^{u}$ &  $C^{u}$ \\
\midrule
100   &     33 &  36 &      48  \\
400   &     38 &   48 &      48 \\
500   &     34 &  54 &       51 \\
800   &     40 &   45 &      47 \\
1K  &     34 &  49 &       47  \\
2K  &      48 &  52 &       54  \\
5K &     28 &   42 &       51  \\
\hline
\hline
 Mean &     36.4 &   46.6 &      49.4  \\
Std  &     \textbf{6.37} &   6.16 &   2.63      \\
\bottomrule
\end{tabular}%
\end{minipage}%
\begin{minipage}{0.45\textwidth}
\begin{tabular}{lc}
\toprule
 TEMI &  CIFAR10  \\
 $\gamma=1$, $C=500$ &  $C^{u}$  \\
\midrule
DINO ViT-B/16 \cite{dino}  &     34 \\
MoCOv3 ViT-B/16 \cite{mocov3}  &     39 \\
iBOT ViT-L/14 \cite{zhou2021ibot}    &     45 \\
OpenCLIP ViT-G/14  \cite{cherti2023reproducible_clipg}  &     47 \\
DINOv2  ViT-g/14  \cite{oquab2023dinov2}   &     50  \\
\hline
\hline
 Mean &     43 \\
Std  &    6.44  \\
\bottomrule
\end{tabular}
\end{minipage}
\label{tab:backbones_beta_1_cifar10}
\end{center}
\end{table*}

\subsection{TEMI with different backbones.}
Here, we report ANMI across various cluster sizes based on the result reported in the main paper (Fig. 5, main paper). For all the conducted experiments, we used TEMI with $\gamma=0.6$. Apart from having roughly the same FID, we can observe the ranking of backbones w.r.t ANMI is not consistent across cluster sizes.

\begin{table}
\centering
\captionof{table}{CIFAR10 ANMI across different cluster sizes and state-of-the-art feature extractors used for TEMI clustering with $\gamma=0.6$. We only reported the ANMI for $C=100$ in the main manuscript.}
\begin{tabular}{lccc}
    \toprule
    TEMI ($\gamma=0.6$) &  ANMI & ANMI & ANMI  \\
    $C$ &  $50$ & $100$ & $200$  \\
                \midrule
                MoCov3 ViT-B/16 \cite{mocov3}  &   65.2 &   59.3  &    55.0  \\
                DINO ViT-B/16 \cite{dino}  &    65.7  &   60.7 &   55.8 \\
                DINOv2 ViT-g/14 \cite{oquab2023dinov2}  &    66.1     &   63.8 &    \textbf{59.3}   \\
                iBOT ViT-L/14 \cite{zhou2021ibot}    &    68.7  &    62.7&    57.4 \\
                CLIP ViT-G/14 \cite{cherti2023reproducible_clipg} &    \textbf{70.6} & \textbf{64.7}  &   58.9  \\
            \bottomrule
        \end{tabular}
        \label{tab:backbones_anmi_cifar10}
\end{table}

\subsection{Dependence on $q(c)$ during generative sampling on balanced classification datasets.}
It is well-established in the clustering literature that $k$-means clusters are highly imbalanced \cite{scan}. To illustrate this in a generative context, we sample from a uniform cluster distribution instead of $q(c)$ for balanced classification datasets (CIFAR10 and CIFAR100). As expected, $k$-means is more dependent to $q(c)$ compared to TEMI, as its FID is significantly deteriorated.

\begin{table*}
\centering
\caption{We report FID for k-means and TEMI with and without considering the training data's cluster distribution $q(c)$. $\mathcal{U}(\{1, .., C \})$ denotes the uniform cluster distribution. We use $C_V=100,200,400$ for CIFAR10, CIFAR100 and FFHQ, respectively. $\Delta$ quantifies the absolute difference.}
\begin{tabular}{l ccc}
\toprule
 {EDM \cite{edm}} & Sampling Distribution & CIFAR10 & CIFAR100 \\
\hline
$k$-means  & $\mathcal{U}(\{1, .., C \})$   & $2.75$ & $2.60$ \\
$k$-means &  $q(c)$  & $1.69$ & $2.21$\\
$\Delta$ ($\downarrow$) & - & $0.79$ & $0.39$ \\ %
\hline
TEMI & $\mathcal{U}(\{1, .., C \})$ & $1.86$ & $2.41$ \\
TEMI & $q(c)$  & $1.67$ & $2.17$ \\
$\Delta$ ($\downarrow$)  & - & \textbf{0.19} & \textbf{0.24}\\
\hline
\end{tabular}

\label{tab:freq-sampling--fids}
\end{table*}

\section{The EDM diffusion baseline.} 
This section briefly summarizes the EDM framework for diffusion models, which was used extensively in this work. For more details and the official EDM code, we refer the reader to the original paper by Karras et al. \cite{edm}.

Given a data distribution $p_{\text{data}}(\rvx)$, consider the conditional distribution $p(\rvx; \sigma)$ of data samples noised with i.i.d.\ Gaussian noise of variance $\sigma^2$. Diffusion-based generative models learn to follow trajectories that connect noisy samples $\rvx \sim p(\rvx; \sigma)$  with data points $\rvy \sim p_{\text{data}}(\rvx)$. Song et al. \cite{SongSDE} introduced the idea of formulating the forward trajectories (from data to noise) using stochastic differential equations (SDE) that evolve samples $\rvx(\sigma)$ according to $p(\rvx; \sigma)$ with $\sigma = \sigma(t)$ as a function of time $t$. They also proposed a corresponding ``probability flow'' ordinary differential equation (ODE), which is fully deterministic and maps the data distribution $p_{\text{data}}(\rvx)$ to the same noise distribution $p(\rvx; \sigma(t))$ as the SDE, for a given time $t$. The ODE continuously adds or removes noise as the sample evolves through time. To formulate the ODE in its simplest form, we need to set a noise schedule $\sigma(t)$ and obtain the \emph{score function} $\nabla_\rvx \log p(\rvx; \sigma)$: 
\begin{equation} \label{eq:ode}
\dd\rvx = -\dot{\sigma}(t) \sigma(t) \nabla_\rvx \log p(\rvx; \sigma) \dd t.
\end{equation}
While mathematical motivations exist for the choice of schedule $\sigma(t)$, empirically motivated choices were shown to be superior \cite{edm}. The main component here, the score function, is learned by a neural network through what is known as \textit{denoising score matching}. The core observation here is that the score does not depend on the intractable normalization constant of $p(\rvx, \sigma(t))$, which is the reason that diffusion models in their current formulation work at all (maybe remove this side-note). Given a denoiser $D(\rvx, \sigma)$ and the L2-denoising error
\begin{equation} \label{eq:L2-obj}
\mathbb{E}_{\rvy \sim p_{\text{data}}} \mathbb{E}_{\rvn \sim \mathcal{N}(0, \sigma^2 I)}[\norm{D(\rvy + \rvn, \sigma) - \rvy}^2],
\end{equation}
we can recover the score function via $\nabla_\rvx \log p(\rvx, \sigma) = (D(\rvx, \sigma) - \rvx) / \sigma^2$.
Thus, parametrizing the denoiser as a neural network and training it on \cref{eq:L2-obj} allows us to learn the score function needed for \cref{eq:ode}. To solve the ODE in \cref{eq:ode}, we can put the recovered score function into \cref{eq:ode} and apply numerical ODE solvers, like Euler's method or Heun's method \cite{heun1998}. The ODE is discretized into a finite number of sampling times $t_0,...,t_N$ and then solved through iteratively computing the score and taking a step with an ODE solver.

\clearpage
\section{Additional implementation details and hyperparameters} \label{app:details}
When searching for $C_V$, we evaluate EDM after training with $M_{img}=100$ and for $M_{img}=200$ once $C_V$ is found. We only report $k$-means cluster conditioning with $k=C_V$. All our reported FID and FDD values are averages over 3 runs of 50k images each, each with different random seeds. Below, we show the hyperparameters we used for all datasets to enable reproducibility. We always use the average FID and FDD for three sets of 50K generated images. The used hyperparameters can be found in \cref{tab:train-hp,temi_hp}
\begin{table*}[b]
\caption{{Hyperparameters used for training EDM and C-EDM. \textbf{Bold} signifies that the value is changing across datasets. All other parameters of the training setup were identical to the specifications of Karras et. al \cite{edm}, which are detailed there.}}
\centering
\label{tab:train-hp} 
\begin{tabular}{lc@{\hskip 5mm}c}
\toprule
Hyperparameter & CIFAR10/CIFAR100 & FFHQ-64/AFHQ-64 \\
\hline
\multicolumn{3}{l}{{{\textbf{Optimization}}}}  \\
optimizer & Adam &  Adam \\
learning rate & 0.001  & 0.001 \\
betas & 0.9, 0.999 & 0.9, 0.999   \\
\textbf{batch size} & 1024 & 512   \\
FP16  &  true & true  \\
\hline
\hline
\multicolumn{3}{l}{{{\textbf{SongUNet}}}}  \\
model channels &  128 & 128  \\
\textbf{channel multiplier}   &  2-2-2 &  1-2-2-2 \\
\textbf{dropout}  &  13\% & 5\% / 25\%  \\
\hline
\hline
\multicolumn{3}{l}{{{\textbf{Augmentation}}}}  \\
augment dim  &  9 & 9   \\
\textbf{probability}  &  12\% & 15\%  \\
\bottomrule
\end{tabular}%
\end{table*}

\begin{table}
\caption{TEMI hyperparameters}
  \centering
    \begin{tabular}{lr}
    \hline
    \textbf{Hyperparameter} & \textbf{Value} \\
    \hline
    \multicolumn{2}{l}{\textbf{Head hyperparameters}} \\
    MLP hidden layers & 2 \\
    hidden dim & 512 \\
    bottleneck dim & 256 \\
    Head final gelu & false \\
    Number of heads (H) & 50 \\
    Loss & TEMI \\
    $\gamma$ & 0.6 \\
    Momentum $\lambda$ & 0.996 \\
    Use batch normalization & false \\
    Dropout & 0.0 \\
    Temperature & 0.1 \\
    Nearest neibohrs (NN) & 50 \\
    Norm last layer & false \\
    \multicolumn{2}{l}{\textbf{Optimization}} \\
    FP16 (mixed precision) & false \\
    Weight decay & 0.0001 \\
    Clip grad & 0 \\
    Batch size & 512 \\
    Epochs & 200 \\
    Learning rate & 0.0001 \\
    Optimizer & AdamW \\
    Drop path rate & 0.1 \\
    Image size & 224 \\
    \hline
    \end{tabular}%
  \label{temi_hp}%
\end{table}

To assign the CLIP pseudo-labels (Sec. 4.4) to the training set, we compute the cosine similarity of the image and label embeddings using openclip's ViT-G/14 \cite{openclip}. The label embeddings use prompt ensembling and use the five prompts: a photo of a $<$label$>$, a blurry photo of a $<$label$>$, a photo of many $<$label$>$, a photo of the large $<$label$>$, and a photo of the small $<$label$>$ as in \cite{ming2022delving}.

\end{document}